\documentclass[10pt,twocolumn,letterpaper]{article}

\usepackage{cvpr}
\usepackage{times}
\usepackage{epsfig}
\usepackage{graphicx}
\usepackage{amsmath}
\usepackage{amssymb}

\usepackage{enumitem}
\usepackage{multirow}                 
\usepackage{subfigure}
\usepackage{array} 
\newcommand{\mb}{\mathbf}
\newcommand{\mc}{\mathcal}
\newcommand{\bs}{\boldsymbol}

\usepackage[pagebackref=true,breaklinks=true,letterpaper=true,colorlinks,bookmarks=false]{hyperref}

\cvprfinalcopy 

\def\cvprPaperID{1841} 

\ifcvprfinal\fi

\begin{document}

\title{Stacked Kernel Network}

\author{Shuai Zhang$^1$, Jianxin Li$^1$, Pengtao Xie$^2$, Yingchun Zhang$^1$, Minglai Shao$^1$, Haoyi Zhou$^1$, Mengyi Yan$^1$\\
$^1$Beihang University \qquad $^2$Carnegie Mellon University\\
{\tt\small \{zhangs,lijx,zhangyc,shaoml,zhouhy,yanmy\}@act.buaa.edu.cn \quad pengtaox@cs.cmu.edu}
}

\maketitle

\vspace{-3mm}
\begin{abstract}
Kernel methods are powerful tools to capture nonlinear patterns behind data. They implicitly learn high (even infinite) dimensional nonlinear features in the Reproducing Kernel Hilbert Space (RKHS) while making the computation tractable by leveraging the kernel trick. Classic kernel methods learn a single layer of nonlinear features, whose representational power may be limited. Motivated by recent success of deep neural networks (DNNs) that learn multi-layer hierarchical representations, we propose a Stacked Kernel Network (SKN) that learns a hierarchy of RKHS-based nonlinear features. SKN interleaves several layers of nonlinear transformations (from a linear space to a RKHS) and linear transformations (from a RKHS to a linear space). Similar to DNNs, a SKN is composed of multiple layers of hidden units, but each parameterized by a RKHS function rather than a finite-dimensional vector. We propose three ways to represent the RKHS functions in SKN: (1)nonparametric representation, (2)parametric representation and (3)random Fourier feature representation.

Furthermore, we expand SKN into CNN architecture called Stacked Kernel Convolutional Network (SKCN). SKCN learning a hierarchy of RKHS-based nonlinear features by convolutional operation with each filter also parameterized by a RKHS function rather than a finite-dimensional matrix in CNN, which is suitable for image inputs. Experiments on various datasets demonstrate the effectiveness of SKN and SKCN, which outperform the competitive methods.
\end{abstract}

\vspace{-4mm}
\section{Introduction}

\begin{figure}
    \centering
    \includegraphics[width=0.42\textwidth]{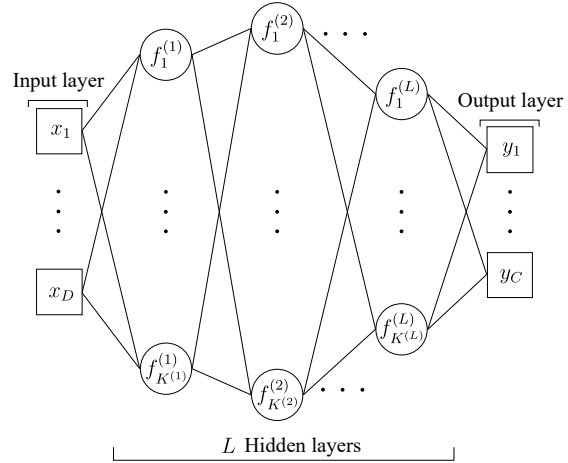}
    \caption{Stacked Kernel Network. A SKN contains multiple hidden layers. Each layer $l$ is associated with a RKHS $\mc{H}^{(l)}$; each unit $k$ in layer $l$ is parameterized by a function $f_k^{(l)}\in \mc{H}^{(l)}$.}
    \label{fig:skn}
    \vspace{-3mm}
\end{figure}

Kernel methods \cite{scholkopf2002learning} \cite{shawe2004kernel} represent a well-established learning paradigm that is able to capture the nonlinear complex patterns underlying data. In kernel methods, the learning is implicitly performed in a high-dimensional (even infinite-dimensional) nonlinear feature space (called Reproducing Kernel Hilbert Space, RKHS) \cite{scholkopf2002learning} via the kernel trick: for a vector $\mb{x}$ in the input space, it is projected into the RKHS using a nonlinear mapping $\phi(\cdot)$, then linear learning is performed on the nonlinear features $\phi(\mb{x})$. $\phi(\cdot)$ is implicitly characterized by a kernel function associated with the RKHS, which known as kernel trick \cite{scholkopf2002learning}.

Classical kernel methods perform single-layer feature learning: the input $\mb{x}$ is transformed into $\phi(\mb{x})$ which are used as the final features on which learning is carried out. Motivated by recent success of deep neural networks which perform hierarchical (``deep") representation learning \cite{bengio2013representation} that is able to capture low-level, middle-level and high-level features, we aim to study ``deep" kernel methods that learn several layers of stacked nonlinear transformations.

Specifically, we propose a Stacked Kernel Network (SKN) that interleaves several layers of nonlinear transformations and linear mappings. Starting from the input $\mb{x}$, a nonlinear feature map $\phi^{(1)}(\cdot)$ is applied to project $\mb{x}$ into a $J^{(1)}$-dimensional ($J^{(1)}$ could be infinite) RKHS $\mathcal{H}^{(1)}$. Then we use a linear mapping $\mb{W}^{(1)}\in\mathbb{R}^{K^{(1)}\times J^{(1)}}$ to project $\phi^{(1)}(\mb{x})$ into a $K^{(1)}$-dimensional linear space $\mathcal{M}^{(1)}$: $\phi^{(1)}(\mb{x})\to \mb{W}^{(1)}\phi^{(1)}(\mb{x})$. Let $\mb{w}_k^\top$ be the $k$-th row vector of $\mb{W}^{(1)}$, according to the definition of RKHS, $\mb{w}_k^\top\phi^{(1)}(\mb{x})$ can be computed as $f_k^{(1)}(\mb{x})$ where $f_k^{(1)}$ is a function in this RKHS. To this end, the representation of $\mb{x}$ in $\mathcal{M}^{(1)}$ can be written as $\mb{h}^{(1)}\in \mathbb{R}^{K^{(1)}}$, where the $k$-th element of $\mb{h}^{(1)}$ is $f_k^{(1)}(\mb{x})$. Then treating $\mb{h}^{(1)}$ as input, we apply the above procedure again: projecting $\mb{h}^{(1)}$ into another RKHS $\mathcal{H}^{(2)}$, followed by a linear projection into another linear space $\mathcal{M}^{(2)}$, getting the representation $\mb{h}^{(2)}$. Repeating this process $L$ times, we obtain a SKN with $L$ hidden layers. SKN contains multiple layers of nonlinear representations which could be infinite-dimensional. This grants SKN vast representation power to capture the complex patterns behind data. On the other hand, after each nonlinear layer, a linear mapping is applied to confine the size of the model, so that the model capacity does not get out of control.

Figure \ref{fig:skn} shows the architecture of SKN. Similar to a deep neural network, it contains multiple hidden layers. The striking difference is in SKN each hidden unit is parameterized by a RKHS function while in DNN the units are parametrized by vectors. The RKHS functions could be infinite-dimensional, which are arguably more expressive than finite-dimensional vectors. As a result, SKN could possess more representational power than DNN. 

Given the multiple layers of RKHS functions in SKN, how to learn them is very challenging. At first, we need to seek explicit representations of these functions. We propose three ways. First, motivated by the representer theorem \cite{scholkopf2002learning}, we parameterize a RKHS function $f$ as a linear combination of kernel functions $k(\cdot,\cdot)$ anchored over training data $\{\mb{x}_i\}_{i=1}^{N}$: $f(\mb{x})=\sum_{i=1}^{N}\alpha_i k(\mb{x}_i,\mb{x})$. Second, we shrink the domain of the functions from the entire RKHS into the image of the nonlinear feature map \cite{scholkopf2002learning}: $\{f_\mb{a}(\mb{x})=k(\mb{a},\mb{x}), \mb{a}\in\mathbb{R}^D\}$, where each function is parameterized by a learnable vector $\mb{a}$.  Third, gaining insight from random Fourier features \cite{rahimi2007random}, a RKHS function $f(\mb{x})$ can be approximated with $\mb{w}^\top z(\mb{x})$ where $z(\mb{x})$ is the random Fourier feature transformation of $\mb{x}$, and $\mb{w}$ is a parameter vector. We use backpropogation algorithm to learn the SKNs under these three representations. Evaluations on various datasets demonstrate the effectiveness of SKN.

The major contributions of this paper are:
\begin{itemize}[noitemsep]
    \item We propose Stacked Kernel Network, which learns multiple layers of RKHS representations.
    \item We study three ways of representing the RKHS functions in SKN to address the computation issue of learning RKHS functions in high-dimensional, even infinite-dimensional.
    \item We design a convolutional architecture of SKN, called Stacked Convolutional Kernel Network (SCKN), to solve visual recognition tasks.
    \item We demonstrate the effectiveness of SKN and SCKN in experiments. 
\end{itemize}
The rest of the paper is organized as follows. Section \ref{sec:method} introduces the Stack Kernel Network and Section \ref{sec:exp} presents experimental results. Section \ref{sec:related work} reviews related works and Section \ref{sec:conclusion} concludes the paper.

\section{Method}
\label{sec:method}
In this section, we introduce the architecture of the Stacked Kernel Network (SKN) and three ways of representing the RKHS functions in SKN. 

\subsection{Kernel methods} \label{sec:kernel}
Kernel methods \cite{scholkopf2002learning} perform learning in a reproducing kernel Hilbert space (RKHS) \cite{scholkopf2002learning} of functions. This RKHS represents a high-dimensional feature space that is more expressive than the input space since it is able to capture the non-linear patterns behind data. The RKHS is associated with a kernel function $k$ and inner product in the RKHS can be computed via evaluating $k$ in the lower-dimensional input space (known as \textit{kernel trick}). Well established kernel methods include support vector machine \cite{burges1998tutorial}, kernel principal component analysis \cite{scholkopf1997kernel}, kernel independent component analysis \cite{bach2002kernel}, Gaussian process \cite{rasmussen2006gaussian}, to name a few.

Kernel methods are featured by three prominent concepts: (1) feature map $\phi$ that maps from a data point $x\in\mathcal{X}$ in the input space to an element of an inner product space (the feature vector); (2) kernel $k(\cdot,\cdot)$ that takes two data points $x,x'\in\mathcal{X}$ and returns a real number; (3) RKHS $\mc{H}$ which is a Hilbert space of functions $f: \mathcal{X}\to \mathbb{R}$. Their relations are as follows. First, feature map defines a kernel. Let $\phi: \mathcal{X}\to \mc{H}$ be a feature map, then $k(x,x')\stackrel{\text{def}}{=}\langle \phi(x),\phi(x')\rangle$ is a kernel. Second, kernel defines feature maps. For every kernel $k$, there exists a Hilbert space $\mc{H}$ and a feature map $\phi: \mc{X}\to \mc{H}$ such that $k(x,x')=\langle \phi(x),\phi(x')\rangle$. Third, RKHS defines a kernel. Every RKHS $\mc{H}$ of functions $f: \mathcal{X}\to \mathbb{R}$ defines a unique kernel $k: \mathcal{X}\times \mathcal{X}\to \mathbb{R}$, called the reproducing kernel of $\mc{H}$. Fourth, kernel defines RKHS. For every kernel $k$, there exists a unique RKHS $\mc{H}$ with reproducing kernel $k$.

\subsection{Deep neural networks} 

We briefly review deep neural networks (DNNs), which inspire us to construct the stacked kernel network. A DNN contains one input layer, several hidden layers and one output layer. Each layer has a set of units. The units between adjacent layers are inter-connected, each connection associated with a weight parameter. To achieve non-linearity, a nonlinear activation function is applied at each hidden unit. Commonly used activation functions include sigmoid, tanh and rectifier linear.

\subsection{Stacked Kernel Network} \label{sec:skn}

Classic kernel methods learn a single layer of nonlinear features, which may not be expressive enough to accommodate complex data. Inspired by the hierarchical representation learning in deep neural networks, we propose Stacked Kernel Network that learns multiple layers of nonlinear features based on RKHS. Figure \ref{fig:skn} shows the architecture of SKN. Similar to a DNN, it consists of an input layer, an output layer and $L$ hidden layers. Each hidden layer is associated with a RKHS and the hidden units therein are parameterized by functions in this RKHS. This is the key difference with DNN where the hidden units are parameterized by weight vectors. Next, we present the detailed construction procedure of SKN. We start with defining the first hidden layer. Let $\mb{x}\in\mathbb{R}^D$ be the input feature vector. We pick up $K^{(1)}$ functions $\{f^{(1)}_i\}_{i=1}^{K^{(1)}}$ from a RKHS $\mathcal{H}^{(1)}$ to map $\mb{x}$ into a $K^{(1)}$-dimensional latent space where the $i$-th dimension is $f^{(1)}_i(\mb{x})$. Then on top of the first hidden layer, we can use $K^{(2)}$ functions $\{f^{(2)}_i\}_{i=1}^{K^{(2)}}$ from another RKHS $\mathcal{H}^{(2)}$ to define the second hidden layer. Repeating this process $L$ times, a SKN with $L$ hidden layers characterized by functions from $L$ RKHS is obtained. The $L$-th hidden layer is utilized to produce the outputs.

\begin{figure}
\centering
\includegraphics[width=0.38\textwidth]{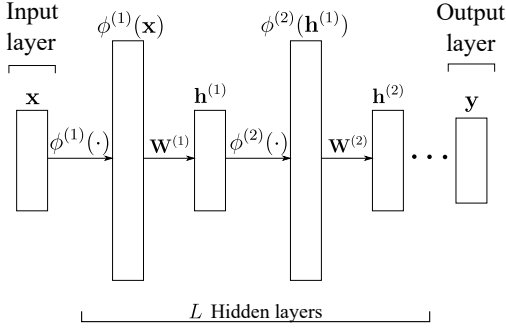}
\caption{Another view of the SKN architecture.}
\label{fig:skn2}
\end{figure}

By stacking multiple RKHS together, the SKN is highly expressive. To better understand this, we present an equivalent architecture in Figure \ref{fig:skn2}. Recall that a RKHS function $f$ is equivalently defined as follows: given an input vector $\mb{x}$, first transform it into $\phi(\mb{x})$ using the reproducing kernel feature map $\phi(\cdot)$, then $f(\mb{x})$ can be define as $\mb{w}^\top\phi(\mb{x})$ where $\mb{w}$ contains linear coefficients. Based upon this definition, the SKN can be represented by interleaving nonlinear projections and linear projections: given the latent representation $\mb{h}^{(l-1)}$ at layer $l-1$, we first transform it into $\phi^{(l)}(\mb{h}^{(l-1)})$ using the nonlinear feature map $\phi^{(l)}(\cdot)$, then use a linear project matrix $\mb{W}^{(l)}$ to map $\phi^{(l)}(\mb{h}^{(l-1)})$ into the latent representation at layer $l$: $\mb{h}^{(l)}=\mb{W}^{(l)}\phi^{(l)}(\mb{h}^{(l-1)})$. The nonlinear features $\phi^{(l)}(\mb{h}^{(l-1)})$ could be infinite-dimensional. A SKN contains $L$ layers of such features that lead to substantial representational power. In between two adjacent layers of nonlinear features, a layer of finite-dimensional linear features $\mb{W}^{(l)}\phi^{(l)}(\mb{h}^{(l-1)})$ is placed. This ensures the size of SKN is properly controlled.

\subsection{Representing RKHS functions in SKN} \label{sec:repre}

A SKN is parameterized with $L$ layers of RKHS functions. While expressive, these functions present great challenges for learning. Unlike weights in DNN that are naturally represented as finite-dimensional vectors, the RKHS functions are of infinite dimension, whose storage and computation are troublesome. To address this issue, we first seek explicit representations of these functions that facilitate learning. We investigate three ways: data-dependent nonparametric representation, data-independent parametric representation and an approximate representation based on random Fourier features.

\subsubsection{Nonparametric representation}

In kernel methods, the most common way to represent a RKHS function is based on the \textit{representer theorem} \cite{scholkopf2001generalized}: given a regularized risk functional $\frac{1}{N}\sum_{i=1}^{N}\ell(f(\mb{x}_i),y_i)+\lambda g(\|f\|_{\mc{H}})$ where $\{(\mb{x}_i,y_i)\}_{i=1}^{N}$ are the training data, $\|f\|_{\mc{H}}$ denotes the Hilbert norm of $f$ and $g(\cdot)$ is an increasing function, the minimizer $f^*$ of this functional admits the following form:
\begin{equation}
f^*(\mb{x})=\sum_{i=1}^{N}\alpha_i k(\mb{x}_i, \mb{x})
\end{equation}
where $k(\cdot,\cdot)$ is a kernel function associated with the RKHS and $\{\alpha_i\}_{i=1}^{N}$ are coefficients. This representation depends on data, hence is referred to as nonparametric representation.

\begin{figure}
\centering
\includegraphics[width=0.28\textwidth]{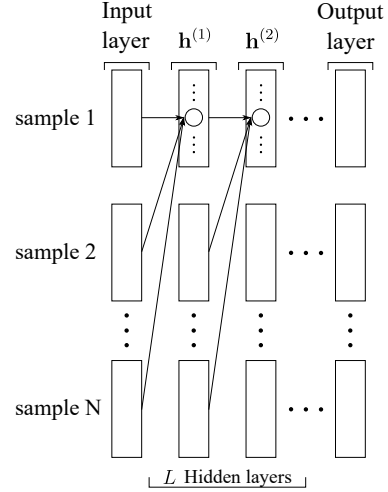}
\caption{Architecture of SKN under nonparametric representation.}
\label{fig:sknnp}
\end{figure}

Specifically, given the hidden states $\{\mb{h}_i^{(l-1)}\}_{i=1}^{N}$ of all training data at layer $l-1$, the RKHS function associated with the $j-$th hidden unit at layer $l$ is defined as:
\begin{equation}
f_j^{(l)}(\mb{x})=\sum_{i=1}^{N}\alpha_i^{(l,j)} k^{(l)}(\mb{h}_i^{(l-1)}, \mb{x})
\end{equation}
where $k^{(l)}(\cdot,\cdot)$ is the kernel function associated with RKHS $\mc{H}^{(l)}$. For data sample $n$, the activation value of hidden unit $j$ at layer $l$ is $f_j^{(l)}(\mb{h}_n^{(l-1)})=\sum_{i=1}^{N}\alpha_i^{(l,j)} k^{(l)}(\mb{h}_i^{(l-1)}, \mb{h}_n^{(l-1)})$. Figure \ref{fig:sknnp} shows the SKN architecture under the nonparametric representation. The input of each hidden unit at layer $l$ are the representations of all samples at layer $l-1$, creating a huge network.

The advantage of the nonparametric representation is that the function set $\{f_j^{(l)}|f_j^{(l)}(\mb{x})=\sum_{i=1}^{N}\alpha_i^{(l,j)} k^{(l)}(\mb{h}_i^{(l-1)}, \mb{x})\}$ contains the global optimal solution, though this optimal may not be achieved due to the non-convexity of SKN. The drawback is the number of parameters $\{\alpha_i^{(l,j)}\}_{i=1}^{N}$ grow linearly with $N$, which is not scalable to large datasets.

\subsubsection{Parametric representation}

\begin{figure}
\centering
\includegraphics[width=0.37\textwidth]{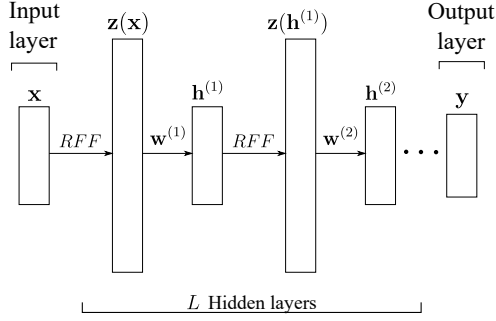}
\caption{Architecture of SKN under random Fourier feature representation.}
\label{fig:sknrff}
\end{figure}

In light of the large computational complexity of the nonparametric representation, we investigate a parametric counterpart. The basic idea is: instead of searching the optimal solution in the entire RKHS, we restrict the learning into a subset of the RKHS where the functions in the subset have nice parametric forms. Specifically, we choose the subset to be the image of the reproducing kernel map:
\begin{equation}
    \{ f|f_{\mathbf{a}}(\mb{x})= k(\mathbf{a},\mb{x}), \mathbf{a} \in \mathbb{R}^D \}
\end{equation}
where $k(\cdot,\cdot)$ is a kernel function associated with the RKHS and $\mathbf{a}$ is the learnable parameter vector which initialized using k-means clustering \cite{celebi2013comparative} from the input samples and trained by gradient decent. Specifically, the RKHS function associated with the $j$-th hidden unit in layer $l$ is:
\begin{equation}
f_j^{(l)}(\mb{x})= k^{(l)}(\mathbf{a}_j^{(l)},\mb{x})
\end{equation} 
where $\mathbf{a}_j^{(l)}$ is the parameter vector of this function. Given the hidden states $\mb{h}^{(l-1)}$ at layer $l-1$, the activation value of this unit is $f_j^{(l)}(\mb{h}^{(l-1)})$. 

The advantage of the parametric representation is that it is independent of data. The disadvantage is the optimal solution of $f$ may not be contained in this subset. Note that if we choose $k(\cdot,\cdot)$ to be the radial basis function (RBF) kernel, then the parametric SKN is specialized to a deep RBF network~\cite{orr1996introduction}. 


\begin{table}[t]\footnotesize
\centering
\begin{tabular}{|c|c|c|c|}
\hline
 & Nonparametric & Parametric & RFF \\
\hline
\hline
Contain \emph{Opt} & Yes  & May not  & No\\
 \hline
 Depend on data & Yes  & No  & No\\
 \hline
 Subset of RKHS & Yes  & Yes  & No\\
 \hline
\# parameters & $O(LKN)$ & $O(LK^2)$  & $O(LKQ)$\\
 \hline
\end{tabular}
\caption{Comparison of three representations ($Opt$ here refers to the global optimal solution).}
\label{table:comp}
\end{table}
\subsubsection{Random Fourier feature representation}

In addition to the nonparametric and parametric representations, we also investigate another approximated representation based on random Fourier features (RFFs) \cite{rahimi2007random}. Given a shift-invariant kernel $k(\mb{x},\mb{y})=k(\mb{x}-\mb{y})$, it can be approximated with RFFs:
\begin{equation}
k(\mb{x},\mb{y})\approx z(\mb{x})^\top z(\mb{y})
\end{equation}
where $z(\mb{x})$ is the RFF transformation. $z(\mb{x})$ is generated in the following way: (1) compute the Fourier transform $p(\bs\omega)$ of the kernel $k(\cdot,\cdot)$; (2) draw $Q$ i.i.d samples $\bs\omega_1,\cdots,\bs\omega_Q\in \mathbb{R}^D$ from $p(\bs\omega)$ and $Q$ i.i.d samples $b_1,\cdots,b_Q\in \mathbb{R}$ from the uniform distribution on $[0,2\pi]$; (3) let $z(\mb{x})=\sqrt{\frac{2}{Q}}[\cos(\bs\omega_1^\top \mb{x}+b_1),\cdots, \cos(\bs\omega_Q^\top \mb{x}+b_Q)]^\top$.

Given a RKHS function $f$, we approximate it using RFF:
\begin{equation}
f(\mb{x})\approx \mb{w}^\top z(\mb{x})
\end{equation}
The RKHS function associated with the $j$-th hidden unit in layer $l$ is:
\begin{equation}
f_j^{(l)}(\mb{x})\approx \mb{w}_j^{(l)}\cdot z(\mb{x})
\end{equation}
where $\mb{w}_j^{(l)}$ is the parameter vector of this function. Given the hidden states $\mb{h}^{(l-1)}$ at layer $l-1$, the activation value of this unit is $\mb{w}_j^{(l)}\cdot z(\mb{h}^{(l-1)})$.

Figure \ref{fig:sknrff} shows the SKN architecture under the RFF representation, where multiple layers of RFF transformations and linear projections are interleaved. The weight parameters in RFF transformation are sampled from the Fourier transform and are fixed during training. The learnable parameters are the weights $\mb{w}$.

\subsubsection{Comparison of three representations}


\begin{figure*}
  \centering
  \subfigure[Nonlinear and linear mapping]{
    \label{fig:skcn1}
    \includegraphics[width=0.42\linewidth]{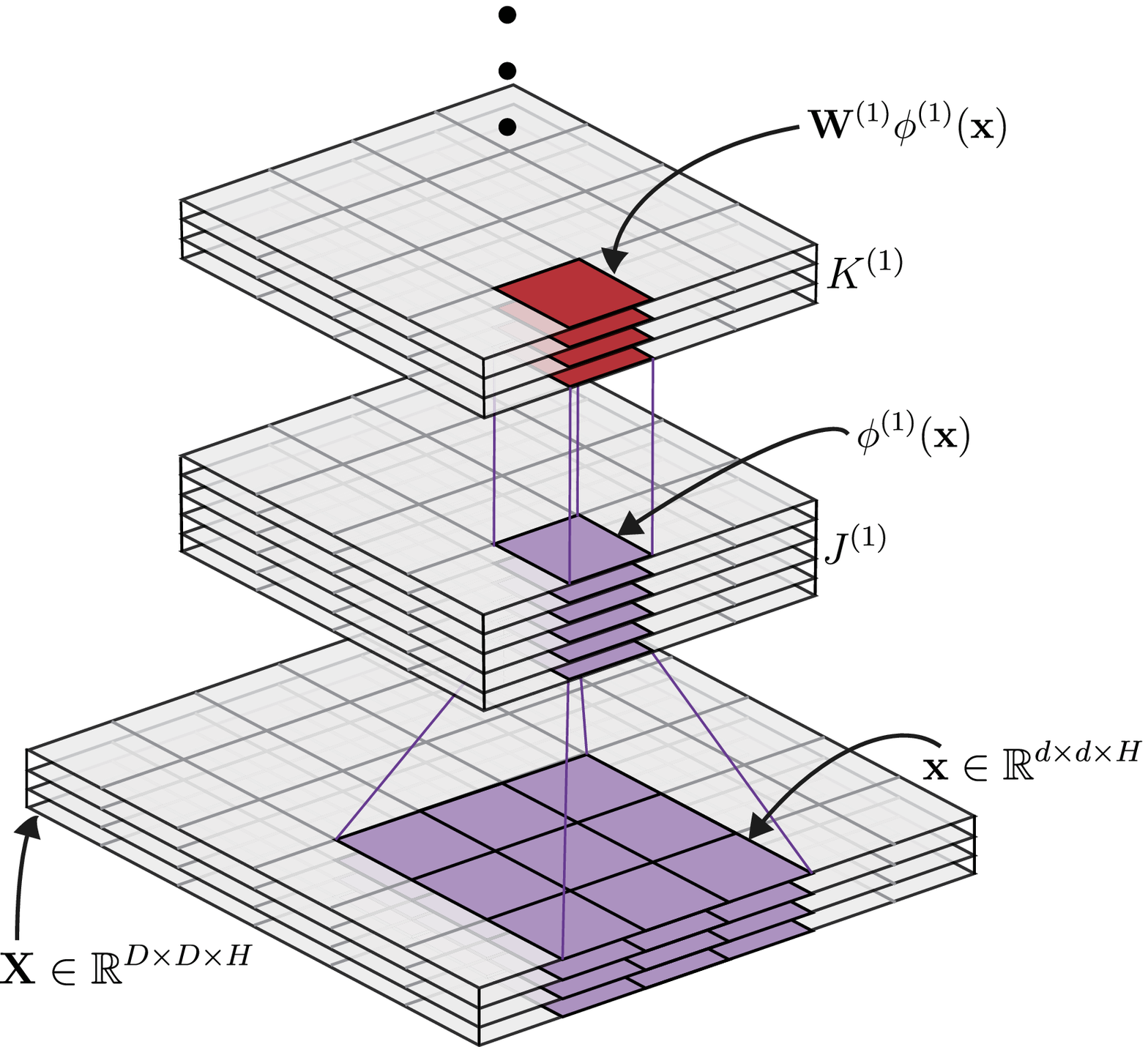}}
  \subfigure[RKHS function mapping]{
    \label{fig:skcn2}
    \includegraphics[width=0.42\linewidth]{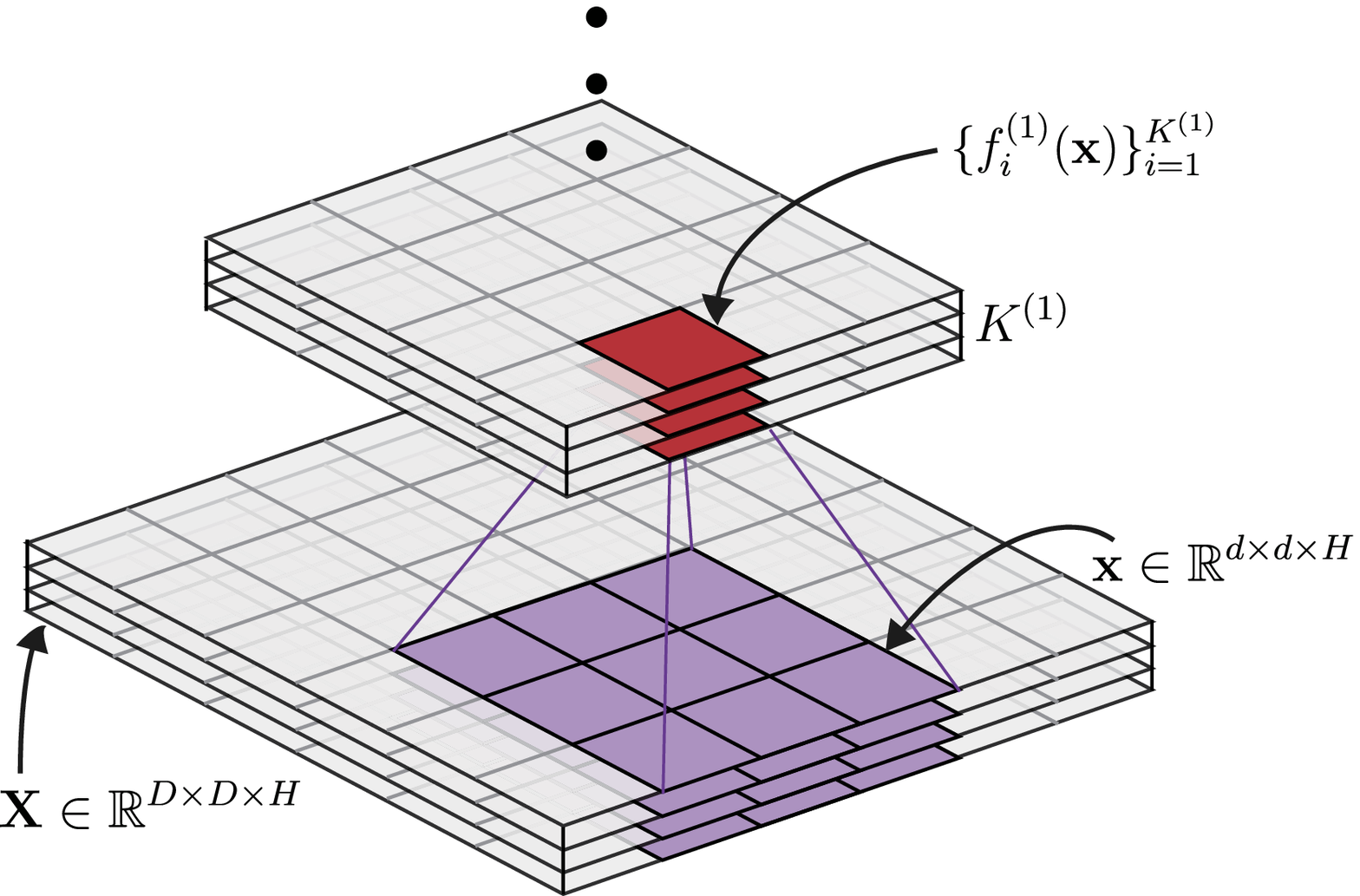}}
  \caption{Two equal way of parameterize the hidden units of Stacked Kernel Convolutional Network. Left: parameterize hidden units by interleave nonlinear and linear mapping. Right: parameterize hidden units by RKHS function mapping.}
\end{figure*}


Table \ref{table:comp} presents a comparison of the three representations. Functions under nonparametric representation contain the global optimal function in the RKHS. Functions under random Fourier feature representation approximate those in the RKHS.  While nonparametric representation contains the global optimal solution in the RKHS, its parameters grow with training data size $N$. Parametric and RFF representations are computationally efficient, however, they have rare chance to reach the global optimal. The number of parameters in all three representations grows with the number of layers $L$ and the number of units $K$ in each layer. For each hidden unit, parametric representation has $K$ parameters and RFF representation has $Q$ (dimension of random features) parameters, which are both much smaller than $N$.

\subsection{Stacked Kernel Convolutional Network} \label{sec:skcnn}

In order to apply SKN to visual recognition problems, we aiming to extend the plain Stacked Kernel Network to a certain advanced architecture which has strong representation ability in visual recognition tasks. Inspired by the recent success of Convolutional Neural Network \cite{lecun1998gradient}, which can extract features patch-wisely from the input feature space by convolution operation, and get strong representation features of the corresponding local area, we design a convolutional architecture for SKN, called Stacked Kernel Convolution Network (SKCN).

Specifically, as shown in Figure \ref{fig:skcn1}, starting from the first kernel convolutional layer, let $\mb{X}\in\mathbb{R}^{D\times D\times H}$ be the input feature space, each patch extracted from $\mb{X}$ is denoted as $\mb{x}\in\mathbb{R}^{d\times d\times H} (d\textless D)$. Similar to SKN, a nonlinear feature map $\phi^{(1)}(\cdot)$ is first applied to project $\mb{x}$ into a $J^{(1)}$-dimensional RKHS $\mathcal{H}^{(1)}$. Then using a linear mapping $\mb{W}^{(1)}\in\mathbb{R}^{K^{(1)}\times J^{(1)}}$ to project $\phi^{(1)}(\mb{x})$ into a $K^{(1)}$-dimensional linear space $\mathcal{M}^{(1)}$: $\phi^{(1)}(\mb{x})\to \mb{W}^{(1)}\phi^{(1)}(\mb{x})$. According to the definition of RKHS, we pick up $K^{(1)}$ functions $\{f^{(1)}_i\}_{i=1}^{K^{(1)}}$ from RKHS $\mathcal{H}^{(1)}$ to replace $\mb{W}^{(1)}\phi^{(1)}(\mb{x})$ to parameterize the hidden units (Figure \ref{fig:skcn2}). To this end, the representation of $\mb{x}$ in $\mathcal{M}^{(1)}$ can be written as $\mb{h}^{(1)}\in \mathbb{R}^{K^{(1)}}$, where the $k$-th element of $\mb{h}^{(1)}$ is $f_k^{(1)}(\mb{x})$. Then on top of the first kernel convolutional layer, we can use $K^{(2)}$ functions $\{f^{(2)}_i\}_{i=1}^{K^{(2)}}$ from another RKHS $\mathcal{H}^{(2)}$ to define the second layer. Repeating this process $L$ times, a SKCN with $L$ kernel convolutional layers characterized by functions from $L$ RKHS is obtained. The $L$-th kernel convolutional layer is utilized to produce the outputs.

In the following experiments, we applied parametric representation and random Fourier feature representation to the SKCN architecture, named P-SKCN and RFF-SKCN respectively.

\begin{table*}[ht]\footnotesize
    \centering
    \begin{tabular}{|c|cccc|cccc|cc|}
    \hline
    \multirow{2}{*}{\textit{Dataset}} & \multicolumn{4}{c|}{DNN (relu)} & \multicolumn{4}{c|}{P-SKN (poly)} & \multicolumn{2}{c|}{NP-SKN (poly)}\\
    \cline{2-11}
        & {1\textit{L}} & {2\textit{L}} & {3\textit{L}} & {4\textit{L}} 
        & {1\textit{L}} & {2\textit{L}} & {3\textit{L}} & {4\textit{L}} 
        & {1\textit{L}} & {2\textit{L}}\\
    \hline
    \hline
    {MNIST}  & 97.19 &\textbf{98.67}  & 98.60 & 98.33 
    & 97.92 & \underline{\textbf{98.71}} & 98.57 & 98.54 
    & 95.83 & \textbf{96.11}\\
    \hline
    {ImageNet-10} & 95.70 &\textbf{97.35} & 97.35 & 97.20 
    & 97.20 & \underline{\textbf{97.65}} & 97.45 & 97.45
    & \textbf{97.25} & 97.20\\
    \hline
    {PenDigits}  & 97.11 & \textbf{98.53} & 98.50 & 98.15
    & 98.68 & \textbf{98.82} & 97.77 & 96.03
    & 95.37 & \textbf{97.88} \\
    \hline
    {SatImage} & 89.85 & \textbf{91.50} & 90.95 & 85.45
    & 91.30 & 91.95 & \underline{\textbf{92.15}} & 92.00
    & 90.00 & \textbf{90.20}\\
    \hline
    {Segment} & 96.86 & 98.25 & \textbf{98.50} & 98.25
    & 97.80 & 98.82 & \underline{\textbf{99.02}} & 98.63
    & 95.47 & \textbf{96.86} \\
    \hline
    {Vowel}  & 52.59 & 57.49 & \textbf{61.75} & 60.25
    & 57.14 & \textbf{63.11} & 59.37 & 57.42
    & 62.34 & \textbf{63.20} \\
    \hline
    \hline
    \multirow{2}{*}{\textit{Dataset}} & \multicolumn{4}{c|}{RFF-SKN (RBF)} & \multicolumn{4}{c|}{P-SKN (RBF)} & \multicolumn{2}{c|}{NP-SKN (RBF)}
    \\
    \cline{2-11}
    & {1\textit{L}} & {2\textit{L}} & {3\textit{L}} & {4\textit{L}} 
    & {1\textit{L}} & {2\textit{L}} & {3\textit{L}} & {4\textit{L}} 
    & {1\textit{L}} & {2\textit{L}}\\ 
    \hline
    \hline
    {MNIST}& 97.97 & 97.99 & 98.08 & \textbf{98.48} 
    & \textbf{98.37} & 98.12 & 97.55 & 97.21 
    & \textbf{96.88} & 96.56 \\
    \hline
    {ImageNet-10}& 97.30 & \textbf{97.35} & 97.20 & 96.35 
    & 97.30 & \textbf{97.55} & 97.45 & 97.35 
    & \textbf{97.20} & 96.90\\
    \hline
    {PenDigits}& 98.20 & 98.04 & 97.84 & \textbf{98.24} 
    & 98.50 & \underline{\textbf{99.06}} & 98.95 & 98.89 
    & 95.81 & \textbf{98.11} \\
    \hline
    {SatImage}& \textbf{90.70} & 90.35 & 90.35 & 87.70 
    & 91.60 & \textbf{92.00} & 90.80 & 89.75 
    & 90.30 & \textbf{90.75}\\
    \hline
    {Segment}& 98.27 & 98.24 & \textbf{98.43} & 98.04
    & 98.83 & \textbf{98.83} & 98.04 & 97.25 
    & 97.25 & \textbf{97.84} \\
    \hline
    {Vowel}& 59.09 & 62.04 & 63.87 & \textbf{65.35}
    & 61.68 & 63.38 & 64.74 & \underline{\textbf{65.43}} 
    & 60.39 & \textbf{61.04} \\ 
    \hline
    \end{tabular}
    \caption{Classification accuracy (\%) on $n$-layer fully-connected network ($n$\textit{L} refers to the number of layers).}
    \label{table:Exp.result.B}
\end{table*}

\section{Experiments}
\label{sec:exp}
In this section, we present experimental results, where we observe (1) ``deep'' SKN/SKCN outperforms single-layer SKN/SKCN; (2) SKN outperforms DNNs and SKCN outperforms CNNs.

\subsection{Datasets}

We compare SKN with DNN on six datasets. \textbf{(1) MNIST} \cite{lecun1998gradient}. It contains images of handwritten digits, represented with 784-dimensional vectors of raw pixels. The training set has 60,000 images and the test set has 10,000 images. \textbf{(2) ImageNet-10}. This dataset contains 10 categories of ImageNet images, each category with 600 images. Images are represented with 128-dimensional convolutional neural network features extracted from the pre-trained ConvNet model \cite{chatfield2014return}. \textbf{(3) PenDigits}. This is a multi-class dataset with 16 integer attributes and 10 classes. It is created by collecting 250 samples from each of 44 writers. \textbf{(4) SatImage}. It is generated from Landsat multi-spectral scanner image data, containing 4435 training samples and 2000 testing samples belonging to 6 classes. Feature dimension is 36. \textbf{(5) Segment}. The data examples were drawn randomly from a database of 7 outdoor image categories. The images were hand-segmented to create a label for every pixel. \textbf{(6) Vowel}. This dataset consists of 11 classes, and each class has 90 10-dimensional samples. 

We compare SKCN with CNN on two datasets. \textbf{(1) MNIST}. It is the same as the MNIST dataset mentioned above. Here we reshape the 784-dimensional vectors into 28$\times$28 images, with 1 grayscale channel. \textbf{(2) CIFAR-10} \cite{krizhevsky2009learning}. It consists of 32$\times$32 images, each with 3 color channels (RGB). The images are from 10 classes. The train and test set contain 50,000 and 10,000 images respectively.

\subsection{Experimental setup}

\begin{figure*}
  \centering
  \subfigure[Accuracy versus iteration of multi-layer CNN on CIFAR-10 dataset.]{
    \label{fig:cnn_acc_iter}
    \includegraphics[width=0.34\linewidth]{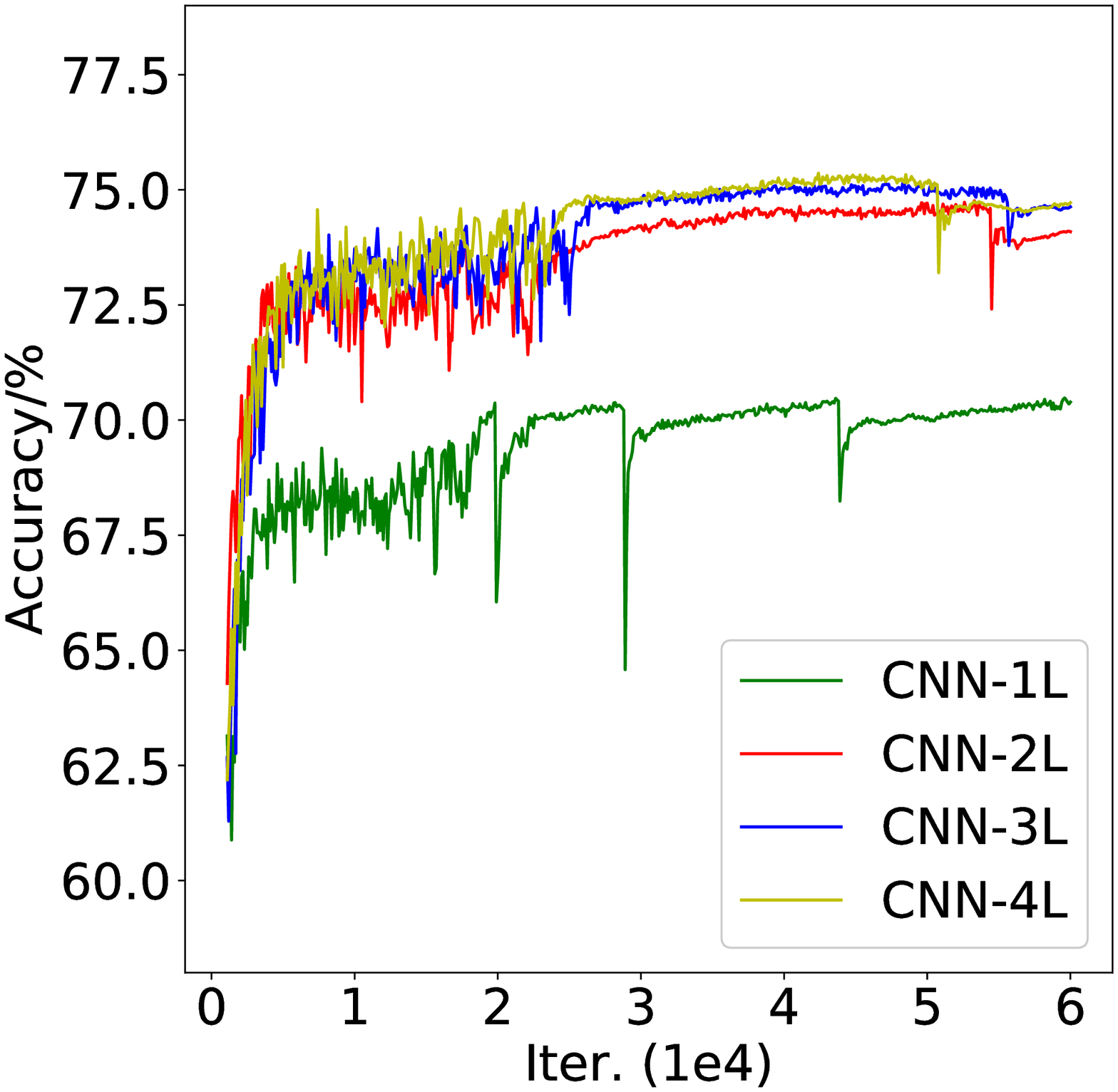}}
  \subfigure[Accuracy versus iteration of multi-layer SKCN on CIFAR-10 dataset.]{
    \label{fig:rff_cnn_acc_iter}
    \includegraphics[width=0.34\linewidth]{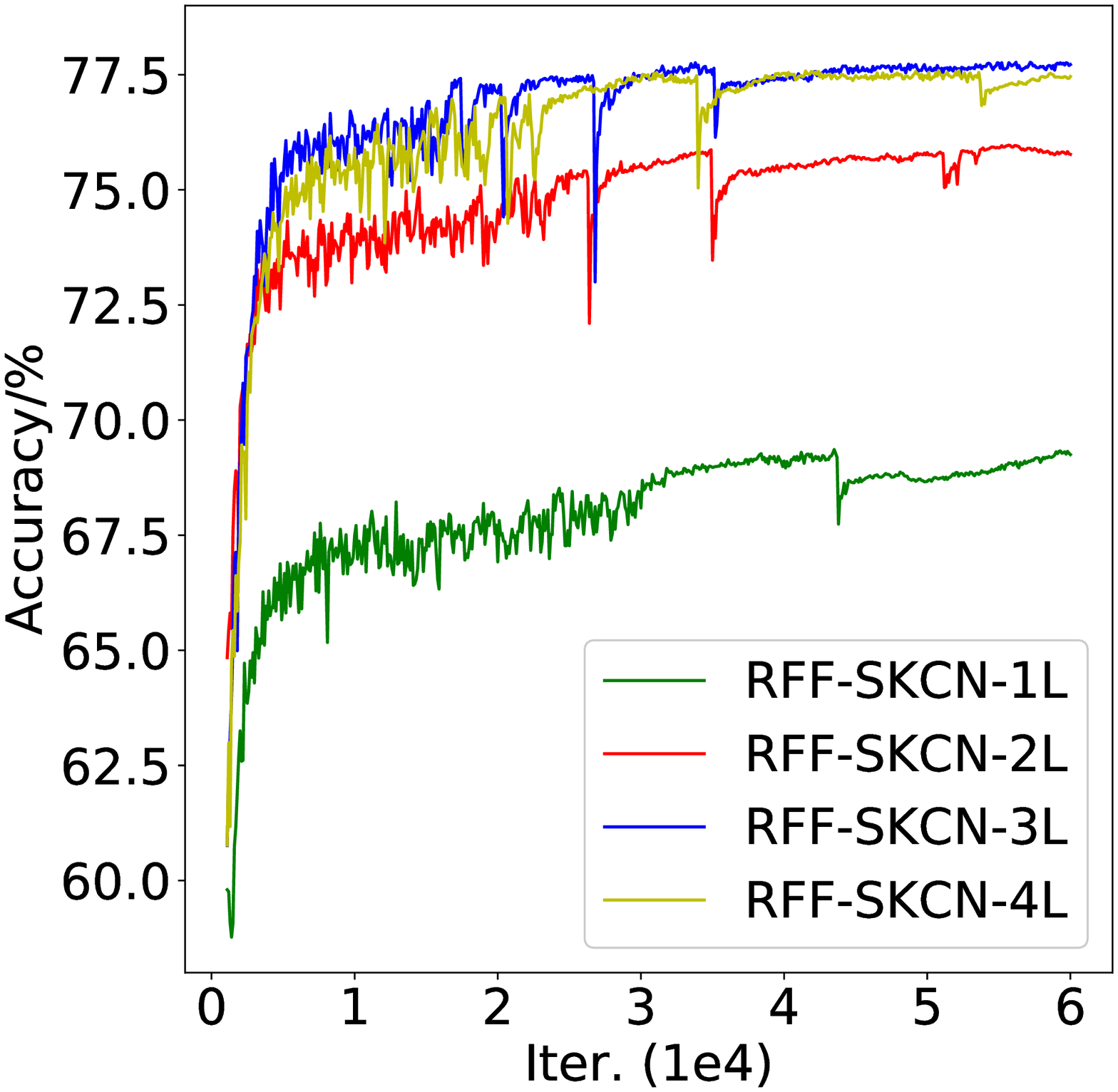}}\\
  \subfigure[Accuracy versus running time of multi-layer CNN on CIFAR-10 dataset.]{
    \label{fig:cnn_acc_time}
    \includegraphics[width=0.34\linewidth]{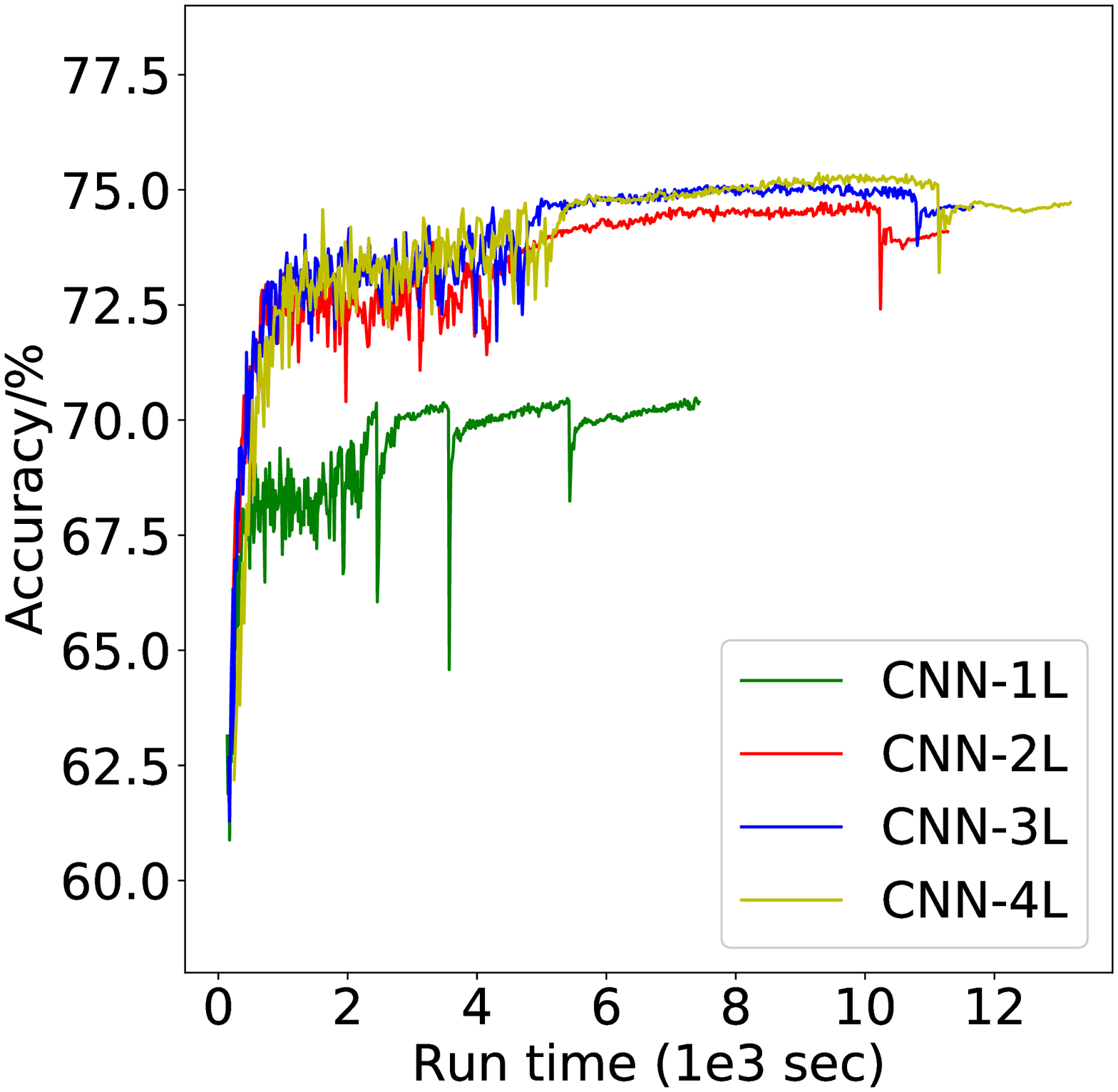}}
  \subfigure[Accuracy versus running time of multi-layer SKCN on CIFAR-10 dataset.]{
    \label{fig:rff_cnn_acc_time}
    \includegraphics[width=0.34\linewidth]{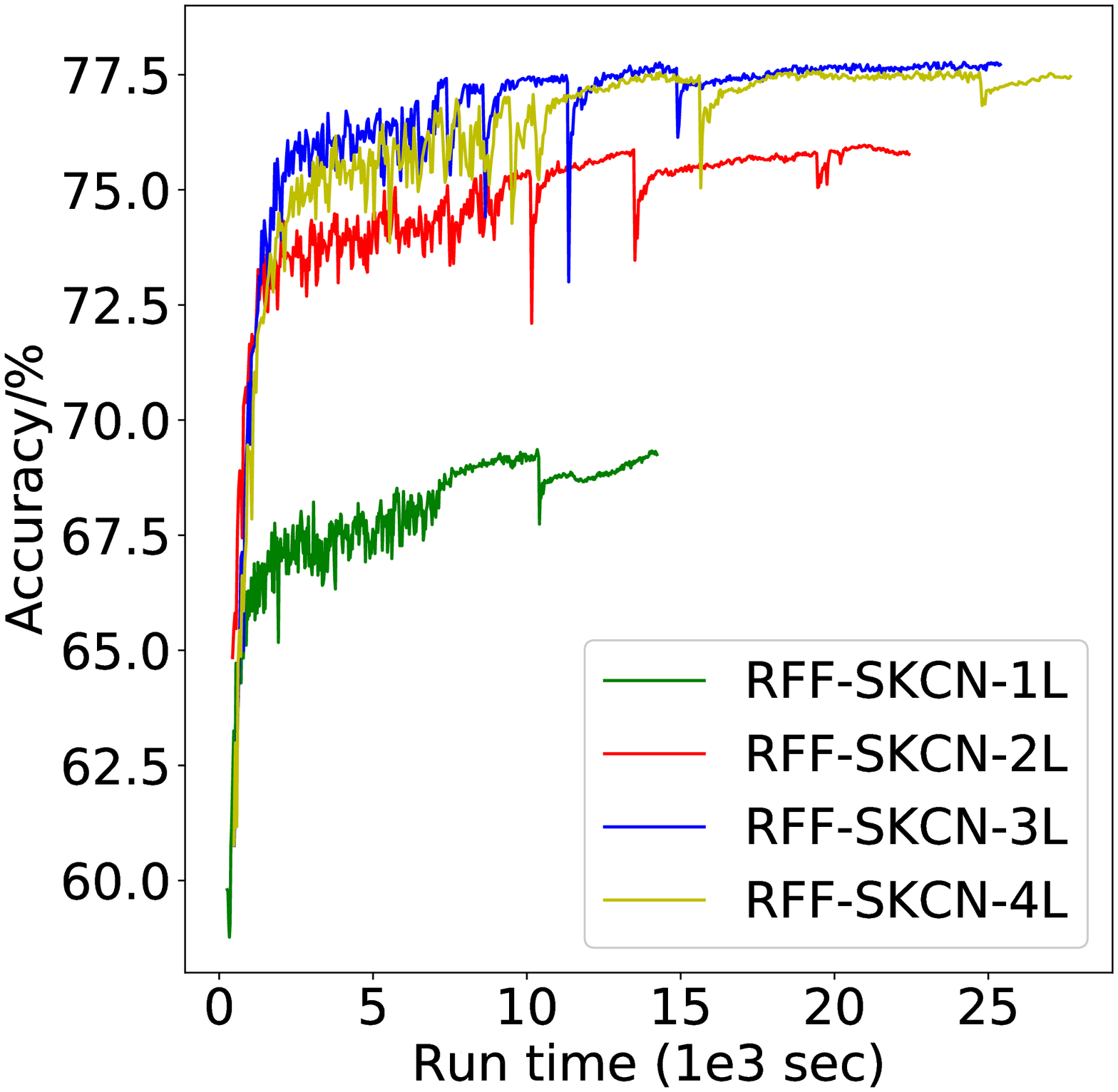}}
  \caption{Accuracy versus iteration/runtime of multi-layer models on CIFAR-10 dataset. (a) and (c): CNN with 1-4 layers; (b) and (d): SKCN with 1-4 layers.}
\end{figure*}

For SKN, we experimented three representations of RKHS function: nonparametric representation (NP-SKN), parametric representation (P-SKN) and random Fourier feature representation (RFF-SKN), and two kernel functions: Radial Basis Function kernel (RBF) $k(x, x') = \exp(\frac{||x-x'||^2}{2\sigma^2})$ and polynomial kernel (poly) $k(x, x') = (x^\top x'+c)^d$. Polynomial kernel is not applicable in RFF-SKN since it is not shift-invariant. The scale parameter $\sigma$ in RBF kernel was tuned in the range $[0.1, 8]$ using 5-fold cross validation. The two parameters $c$ and $d$ in polynomial kernel were tuned in the range $[0, 5]$ (with intervals of 0.5) and $[1, 4]$ (with intervals of 1). We compare with deep neural networks where the activation function is set to rectified linear (relu) with the same hyperparameters all the time. SKN was trained by stochastic gradient decent using cross-entropy loss. The batch size was set to 128, and the learning rate was set to $0.001$. For RFF-SKN, the number of random Fourier feature is set to 5000. We implemented SKN using TensorFlow \cite{tensorflow2015-whitepaper}. These experiments were performed on Linux machines with 32 $\times$ 4.0GHz CPU cores and 256GB RAM.

For SKCN, we experimented two representations of RKHS function: parametric representation (P-SKCN) and random Fourier feature representation (RFF-SKCN). For P-SKCN, we use polynomial kernel function. And for RFF-SKCN, we use RBF kernel function. To get a trade-off between runtime and accuracy, we set the sample number in RFF-SKCN to 2000. The parameters $c$ and $d$ in polynomial kernel and the parameter $\sigma$ in RBF kernel were tuned in same way as them in P-SKN. We compare our methods with CNN using the same hyperparameter setting all the time. For instance, in Table \ref{table:statistics_cnn}, we set both CNN and SKCN (1) Network architecture: conv1-pool1-norm1(layer1), conv2-norm2-pool2(layer2), conv3-norm3-pool3(layer3), conv4-norm4-pool4(layer4); (2) Kernel size: 5$\times$5(layer1), 5$\times$5(layer2), 3$\times$3(layer3), 3$\times$3(layer4); (3) Strides: set 1 for all layers; (5) Padding: Same-padding; (6) Pooling: 2$\times$2 Max-pooling; (4) Dropout \cite{wan2013regularization}: with 50\% or without; (5) Data augmentation: None. The only different setting between CNN and RFF-SKCN are activation functions: CNN (relu), RFF-SKCN (none). In the contrast experiment, we use stochastic gradient decent to minimize the cross-entropy loss, the batch size was set to 128, and we use adaptive learning rate which initialized at $0.001$ and staircase like decayed $10\%$ in every $1000$ iterations. We use local response normalization provided in Tensorflow to normalize the outputs of each layer. These experiments were performed on Linux machines with Tesla K80 GPUs and 256GB RAM.

\subsection{Results}

Table \ref{table:Exp.result.B} shows the classification accuracy of DNN and SKN with different representation of RKHS functions and kernel functions. The accuracy of NP-SKN with more than 2 layers are not available since the model is too large and it takes too much time to converge. From this table, we make the following observations. (1) P-SKN (with polynomial kernel or RBF kernel) outperforms DNN. For instance, on Vowel, P-SKN with RBF kernel achieves an accuracy of 65.43\% while the accuracy achieved by DNN is 61.75\%. The hidden units in SKN are parameterized by infinite-dimensional RKHS functions, which are more expressive than finite-dimensional weight vectors in DNN. Thus SKN is more capable to capture the complex patterns behind data and achieves higher classification accuracy. (2) In most cases, SKN with $\geq 2$ hidden layers outperforms that with one single hidden layer. For example, on the Segment dataset, P-SKN (poly) with 3 layers achieve an accuracy of 99.02\% while that with 1 layer achieves 97.80\%. This demonstrates that ``deep" SKN is better than single-layer kernel method. Stacking multiple layers of RKHS functions together improves the representation power of kernel methods and results in better performance. However, the number of layers cannot be too large, which otherwise leads to overfitting and hurts generalization performance on the test set. For instance, on the SatImage dataset, P-SKN (RBF) with 4 layers performs worse than that with 2 layers. (3) The performance of P-SKN is better than NP-SKN and RFF-SKN. For NP-SKN, its number of parameters grows linearly with data size, which easily leads to overfitting. For RFF-SKN, the RFF-represented functions are not true functions in the RKHS, but rather approximations, which suffers an approximation error (but overall, its performance is very close to P-SKN). In contrast, the functions with parametric representations are exactly from the RKHS and their parameters do not depend on data size.

\begin{table*}[t]\footnotesize
\centering
{\begin{tabular}{|c|cccc|cccc|cccc|}
\hline
\multirow{2}{*}{\textit{Dataset}} & \multicolumn{4}{c|}{CNN} & \multicolumn{4}{c|}{RFF-SKCN} & \multicolumn{4}{c|}{P-SKCN}\\
\cline{2-13} & {1\textit{L}} & {2\textit{L}} & {3\textit{L}} & {4\textit{L}}
& {1\textit{L}} & {2\textit{L}} & {3\textit{L}} & {4\textit{L}}
& {1\textit{L}} & {2\textit{L}} & {3\textit{L}} & {4\textit{L}}\\
\hline
\hline
{MNIST}  & 99.26 & 99.47 & 99.43 & \textbf{99.51} 
& 99.31 & \underline{\textbf{99.63}} & 99.62 & 99.58 
& 99.30 & \textbf{99.49} & 99.27 & 99.27\\
\hline
{MNIST (\textit{drop})} & 99.38 & 99.54 & 99.50 & \textbf{99.56} 
& 99.38 & 99.60 & \underline{\textbf{99.63}} & 99.58
& 99.10 & 99.40 & 99.37 & \textbf{99.41} \\
\hline
{CIFAR-10} & 70.03 & 72.83 & 73.24 & \textbf{74.86}
& 69.36 & 75.97 & \underline{\textbf{77.77}} & 77.59 
& 67.77 & 73.33 & \textbf{73.41} & 72.19\\
\hline
{CIFAR-10 (\textit{drop})} & 71.21 & 75.60 & 74.79 & \textbf{75.88}
& 72.45 & 76.82 & \underline{\textbf{79.06}} & 77.13 
& 67.84 & 74.71 & \textbf{76.79} & 75.31 \\
\hline
\end{tabular}}
\caption{Classification accuracy (\%) of CNN and SKCN.}
\label{table:statistics_cnn}
\end{table*}

\begin{figure}[t]
    \centering
    \includegraphics[width=0.3\textwidth]{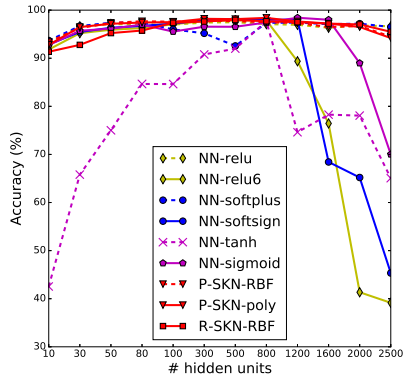}
    \caption{Accuracy versus the number of hidden units.}
    \label{fig:node}
\end{figure}

Besides, we design a experiment to find out the sensitiveness of SKN and DNNs. Figure \ref{fig:node} shows how accuracy changes w.r.t. the number of hidden units $K$, on the MNIST dataset. For DNN, we experimented different acivation functions including relu\cite{nair2010rectified}, relu6\cite{krizhevsky2010convolutional}, softplus\cite{glorot2011deep}, softsign\cite{bergstra2009quadratic}, tanh and sigmoid. As can be seen, overall, SKNs are not sensitive to $K$. The accuracy remains stable as $K$ increases from 10 to 2500. In contrast, DNNs are very sensitive to $K$. For each method, the best accuracy is achieved at a $K$ that is in the middle ground. A smaller $K$ is not expressive enough and a larger $K$ leads to overfitting.

Table \ref{table:statistics_cnn} listed experiment results of CNN and SKCN. We have two major observation from this table: (1) RFF-SKCN outperforms CNN. SKCN achieve higher accuracy on most dataset with same layer number. For instance, RFF-SKCN achieves 79.06\% with dropout and 77.77\% without dropout on CIFAR-10 dataset with 3-layer, while CNN only get 74.79\% and 73.24\% respectively. (2) In most cases, SKCN with $\geq 2$ layers outperforms that with one single layer.

Next, We show the detail performance of CNN and RFF-SKCN with different layer on CIFAR-10 dataset from Figure \ref{fig:cnn_acc_iter} to Figure \ref{fig:rff_cnn_acc_time}. First, from Figure \ref{fig:cnn_acc_iter} and Figure \ref{fig:rff_cnn_acc_iter} we can observe that RFF-SKCN exhibits considerably higher testing accuracy and is more generalizable than the baseline architecture CNN, which indicates that the infinite-dimensional RKHS functions are more expressive than finite-dimensional filters in CNN. Second, although RFF-SKCN outperforms CNN, from Figure \ref{fig:cnn_acc_time} and Figure \ref{fig:rff_cnn_acc_time} we can observe that our method RFF-SKCN take nearly twice as long to run a single iteration. It is mainly because RFF representation need to train a larger $W$ than CNN, and as a result of this, the learned parameters in RFF-SKCN can be more expressive than CNN.

\section{Related work}
\label{sec:related work}
Several studies have been performed to bridge kernel methods and deep learning. \cite{cho2009kernel} aim to define an ``deep" kernel $k$ by first successively composing the same nonlinear transformation $\phi$ multiple times over the input $x, x'$: $\phi(\phi(\cdots\phi(x)))$, then defining $k$ as $k(x,x')=\phi(\phi(\cdots\phi(x))) \cdot \phi(\phi(\cdots\phi(x')))$. Using the kernel trick, these $\phi$ can be replaced with kernel functions. However, if the representation $\phi(\phi(\cdots\phi(x)))$ is not used for defining kernel, but rather making predictions (which is the case in our work), it is unclear how to deal with these $\phi$. \cite{varma2013local} propose local deep kernel learning where the local feature space are represented with a hierarchy of nonlinear transformations guided by a tree structure. \cite{wilson2015deep} use deep neural network (DNN) to define a kernel function $k$. Given two input data $x$ and $x'$, they are first fed into a DNN $g$, generating latent representations $g(x)$ and $g(x')$, which are then fed into a kernel function $\hat{k}$. Overall, $k$ is defined as $\hat{k}(g(x),g(x'))$. The major difference between these works with ours is that they utilize DNN or nested nonlinear transformations to define a single kernel function, while our work defines a network with multiple layers of RKHS functions. \cite{damianou2013deep} propose a generative model which stacks multiple layers of Gaussian processes while our work is a discriminative model that stacks multiple layers of RKHS functions. \cite{mairal2014convolutional} use convolutional neural network to approximate the kernel map while our work uses random Fourier features as building blocks of deep networks.

\section{Conclusion}
\label{sec:conclusion}
In this paper, we propose a ``deep" kernel method -- Stacked Kernel Network, that learns a hierarchy of RKHS functions. SKN consists of multiple layers of interleaving nonlinear and linear projections. The nonlinear projection is carried out by the reproducing kernel feature map associated with a RKHS and the resultant features could be infinite-dimensional. A SKN is equipped with multiple such feature maps that bring in high representation power. To avoid the model size of SKN out of control, immediately after each nonlinear transformation, a linear projection is applied to map the infinite-dimensional nonlinear space to a finite-dimensional linear space. In the end, a SKN is composed of multiple hidden layers, each associated with a RKHS and each unit therein is parameterized by a function in that RKHS. We investigate three ways to represent the RKHS functions to make their learning tractable: data-dependent nonparametric representation based on the representer theorem, data-independent parametric representation and an approximate representation based on random Fourier features. Experiments on various datasets demonstrate the effectiveness of SKN.

{\small
\bibliographystyle{ieee}
\bibliography{paper}
}

\end{document}